\documentclass[10pt,twocolumn,letterpaper]{article}

\makeatletter
\@namedef{ver@everyshi.sty}{}
\makeatother
\usepackage[dvipsnames]{xcolor}
\usepackage{tikz}
\usetikzlibrary{positioning}

\usepackage{cvpr}              
\usepackage{adjustbox}
\usepackage{amsmath}
\usepackage{amssymb}
\usepackage{booktabs}

\usepackage{bm}
\usepackage{multirow}
\usepackage{subcaption}
\usepackage{algorithm}
\usepackage{algpseudocode}
\usepackage{float}
\usepackage{capt-of}
\usepackage{lipsum}
\usepackage{bbm}
\usepackage{colortbl}

\usepackage{amsmath}
\usepackage{pifont}

\definecolor{cvprblue}{rgb}{0.21,0.49,0.74}
\usepackage[pagebackref,breaklinks,colorlinks,allcolors=cvprblue]{hyperref}

\definecolor{waymogreen}{rgb}{0,0.91,0.62}

\definecolor{waymoblue}{rgb}{0,0.47,1}

\newcommand{\ours}{SceneDiffuser++}
\newcommand{\task}{CitySim}

\newcommand{\boldparagraph}[1]{\vspace{0.1cm}\noindent{\bf #1} }

\colorlet{tableheadcolor}{gray!25} 
\colorlet{tablerowcolor}{gray!10} 
\newcommand{\rowcolorize}{\rowcolor{tablerowcolor}} %

\colorlet{tablebluerowcolor}{blue!10} 

\newcommand{\Tau}{\mathcal{T}}

\def\paperID{17198} 
\def\confName{CVPR}
\def\confYear{2025}

\begin{document}

\title{\ours: City-Scale Traffic Simulation via a Generative World Model}

\author{
Shuhan Tan$^{2*}$ \quad John Lambert$^{1}$ \quad Hong Jeon$^{1}$ \quad 
Sakshum Kulshrestha$^{1}$ \quad
Yijing Bai$^{1}$ \\
Jing Luo$^{1}$  \quad
Dragomir Anguelov$^{1}$ \quad
Mingxing Tan$^{1}$ \quad
Chiyu Max Jiang$^{1}$\\
$^{1}$Waymo LLC \quad $^{2}$UT Austin
}


\setlength{\abovedisplayskip}{0.5\abovedisplayskip}
\setlength{\belowdisplayskip}{0.5\belowdisplayskip}
\setlength{\abovecaptionskip}{0.5\abovecaptionskip}
\setlength{\belowcaptionskip}{0.5\belowcaptionskip}

\twocolumn[{%
\renewcommand\twocolumn[1][]{#1}%
\maketitle

\vspace{-3em}
\begin{center}
    \centering
    \begin{tikzpicture}[node distance=0.6cm, every node/.style={inner sep=0pt}]
    \newcommand{\imagesize}{0.21\textwidth} 
    \newcommand{\margin}{2px} 

    \node (goal) {\includegraphics[trim={\margin, \margin, \margin, \margin},clip,width=\imagesize, height=\imagesize]{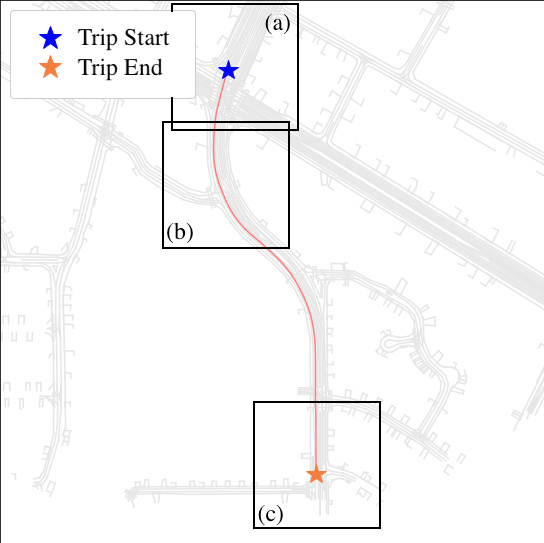}};
    \node[right=of goal] (initial) {\fbox{\includegraphics[trim={\margin, \margin, \margin, \margin},clip,width=\imagesize, height=\imagesize]{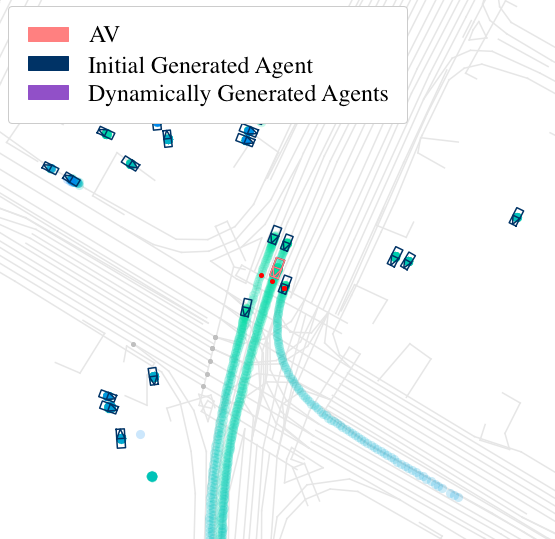}}};
    \node[right=of initial] (rollout) {\fbox{\includegraphics[trim={\margin, \margin, \margin, \margin},clip,width=\imagesize, height=\imagesize]{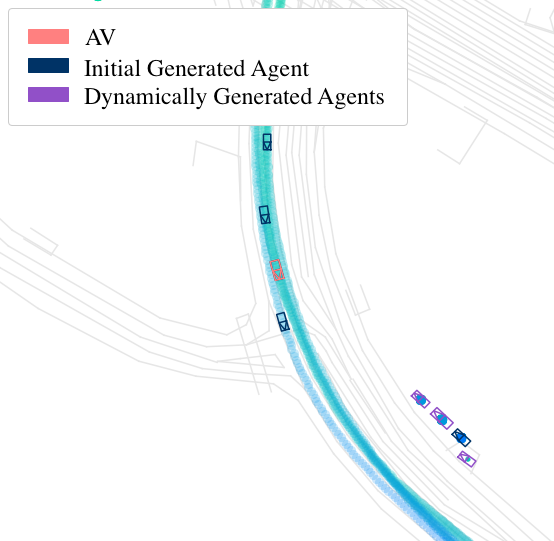}}};
    \node[right=of rollout] (final) {\fbox{\includegraphics[trim={\margin, \margin, \margin, \margin},clip,width=\imagesize, height=\imagesize]{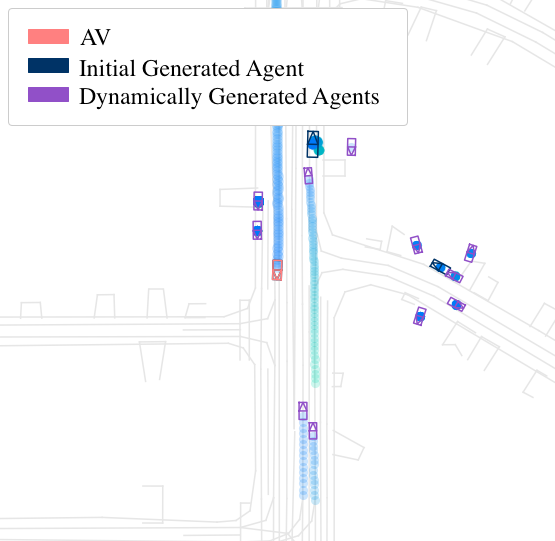}}};

    \draw[->, shorten >=5pt, shorten <=5pt] (goal) -- (initial);
    \draw[->, shorten >=5pt, shorten <=5pt] (initial) -- (rollout);
    \draw[->, shorten >=5pt, shorten <=5pt] (rollout) -- (final);

    \node[below=0.5cm of goal] {Simulation Route};
    \node[below=0.5cm of initial] {(a) Initial Scene Generation};
    \node[below=0.5cm of rollout] {(b) Long Simulation Rollout};
    \node[below=0.5cm of final] {(c) Final Destination};
\end{tikzpicture}

    \vspace{-5mm}
    \captionof{figure}{Overview. \ours{} is a single unified, end-to-end trained generative world model that enables \task{}: city-scale traffic simulation that takes in a large map region, start and end points, and simulates everything in between, from initial scene generation, agent behavior prediction, occlusion reasoning, dynamic agent generation (spawning and removal) to environment simulation (traffic lights).}
    \label{fig:teaser}
\end{center}%

}]

%
\begin{abstract}
The goal of traffic simulation is to augment a potentially limited amount of manually-driven miles that is available for testing and validation, with a much larger amount of simulated synthetic miles. The culmination of this vision would be a generative simulated city, where given a map of the city and an autonomous vehicle (AV) software stack, the simulator can seamlessly simulate the trip from point A to point B by populating the city around the AV and controlling all aspects of the scene, from animating the dynamic agents (e.g., vehicles, pedestrians) to controlling the traffic light states. We refer to this vision as \emph{\task{}}, which requires an agglomeration of simulation technologies: scene generation to populate the initial scene, agent behavior modeling to animate the scene, occlusion reasoning, dynamic scene generation to seamlessly spawn and remove agents, and environment simulation for factors such as traffic lights. While some key technologies have been separately studied in various works, others such as dynamic scene generation and environment simulation have received less attention in the research community. We propose SceneDiffuser++, the first end-to-end generative world model trained on a single loss function capable of point A-to-B simulation on a city scale integrating all the requirements above. We demonstrate the city-scale traffic simulation capability of SceneDiffuser++ and study its superior realism under long simulation conditions.
We evaluate the simulation quality on an augmented version of the Waymo Open Motion Dataset (WOMD) with larger map regions to support trip-level simulation.
\begingroup
\renewcommand\thefootnote{*}
\footnotetext{Work done as an intern at Waymo.}
\endgroup
\end{abstract}

\vspace{-3mm}
\section{Introduction}

Imagine an ideal traffic simulation at the city-scale: Starting from a logged or synthetic scene, we initiate the simulation. The virtual world comes alive with agents behaving realistically: cars navigate roads, pedestrians cross streets, and interactions unfold naturally.  A pedestrian emerges from behind a bus, prompting a reaction from the ego agent. Vehicles disappear and reappear as they become occluded and disoccluded. Turning onto a new road reveals a fresh stream of traffic. The ego vehicle responds to traffic signals, stopping at red lights and proceeding when they turn green. This simulation persists for a long duration, allowing trip-level evaluations of driving by generating a dynamically populated virtual city with continuous agent interactions.


We refer to such a city-scale closed-loop traffic simulation system as \task{} (See Fig.~\ref{fig:teaser}). \task{} can enable point-to-point driving simulation for obtaining trip-level statistics. This allows holistic driving assessment, for instance trip-level travel time comparisons to average human drivers, pick-up and drop-off quality assessment, as well as evaluation of driving behaviors during the trip, including safety and driving quality. Such simulators also allow for playing out events that take longer to unfold, such as interactions between an AV and emergency vehicles. They can also facilitate pre-release evaluation of AV software by estimating safety and quality related rates, system-level hillclimbing, and system-level fault discovery\cite{Corso21jair_SurveyBlackBoxSafetyValidation, Agha18tomacs_StatisticalModelChecking}. CitySim systems stand in contrast to simulation frameworks based on simulating logged events (usually $<$ 10s), which is the mainstream setup in most existing frameworks \cite{Montali23neurips_wosac, Gulino23neurips_Waymax}.

Transitioning from event-level simulation to trip-level simulation requires a step-function improvement in simulation capabilities. While event-level simulations are short in duration, naively extending them to longer durations triggers a host of realism issues. In longer simulation, the initial logged agents \cite{Chang19cvpr_Argoverse, Ettinger21iccv_WOMD, Wilson21neurips_Argoverse2, Caesar20cvpr_nuScenes} might leave the periphery of the AV while new agents might continuously and seamlessly appear, mandating dynamic agent generation to handle agent spawning and removal. The need for dynamic agent generation is more critical in cases where the AV takes a different route or speed profile which might result in the AV very quickly turning into an empty street. Furthermore, as the simulated AV heads into regions of the map not traversed in the initial log, traffic light states and other environment factors need to be simulated as well.  Simulation artifacts arising from high pose divergence between the logged and simulated AV are referred to as ``simulation drift'' \cite{Bergamini21icra_SimNet}. 



These unrealistic behaviors in high pose divergence scenarios highlight three critical, yet often overlooked, capabilities in learned traffic simulation:  dynamic agent generation (including \emph{agent spawning} for new agents entering the scene, \emph{agent removal}  for agents exiting the scene), \emph{occlusion reasoning}, and the dynamic handling of \emph{critical environmental factors} like traffic lights. To our knowledge, most of the aforementioned technologies are not investigated in existing learned simulation models.

In this work, we bring together this vision of a realistic and dynamically populated virtual city that enables trip-level simulation in a single end-to-end learned, generative world model that we refer to as \ours{}. \ours{} is a diffusion model that is solely trained on the diffusion denoising objective, yet supports all aforementioned capabilities via simple autoregressive rollout. Following \citet{Jiang24neurips_scenediffuser}, we model the problem as denoising the scene-tensor, with various key insights. First, we observe that agent spawning, removal and occlusion reasoning can be jointly modeled simply via predicting an additional validity (or equivalently, visibility) channel along with other agent features such as $x, y$, size, type, etc. Though conceptually simple, this requires diffusion to learn to generate sparse tensors without prespecified sparse structure. We propose a simple yet effective training loss formulation and inference-time diffusion sampler modification to allow stable training and sampling of such models. Finally, we propose a novel architecture change that allows simulating the joint rollouts of various non-homogeneous scene elements (e.g., agents and traffic lights with different feature sizes). We propose novel ways to evaluate the realism of such trip-level simulation, and benchmark and ablate our design choices on a version of the Waymo Open Motion Dataset (WOMD) augmented with enlarged kilometer-scale map regions for long rollouts. 


In summary, our contributions are as follows:

\begin{itemize}
    \item We conceptualize the novel city-scale traffic simulation task: \task{}, which focuses on trip-level simulations.
    \item In contrast to event-level simulations, we identify novel challenges from trip-level simulations, and propose novel evaluation metrics for evaluating the realism of agent spawning, removal, occlusion and traffic light simulation.
    \item We propose a unified generative world model: \ours{}, enabling realistic long simulations while accounting for dynamic agent generation, occlusion reasoning and traffic light simulation via simple autoregressive rollout using a novel method to generate sparse tensors.
    \item We demonstrate our performance on a map-augmented WOMD dataset \cite{Ettinger21iccv_WOMD} and achieve state-of-the-art trip-level simulation realism.
\end{itemize}

 \begin{figure*}[t]
  \centering
    \includegraphics[trim={0 0.25cm 0 0cm},clip,width=\linewidth]{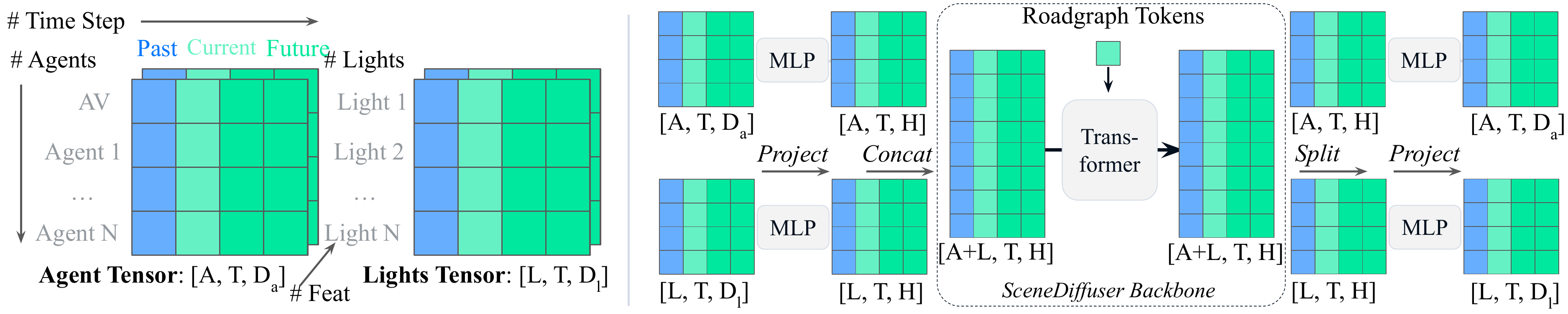}
    \caption{Multi-Tensor Diffusion jointly denoises the set of scene elements in a joint fashion. \textbf{Left}: the two scene tensors for agents and traffic lights with a varying number of elements and feature dimensions. \textbf{Right}: denoiser architecture for multi-tensors that first projects and homogenizes different scene tensors to the same number of latents before concatenating into a full scene tensor to pass through an axial-attention based transformer backbone \cite{Jiang24neurips_scenediffuser}.}
    \label{fig:multitensor}
    \vspace{-3mm}
\end{figure*}

\section{Related Work}

\boldparagraph{World Models} Recently, \emph{world models}, i.e. simulators of how the physical world evolves over time, have attracted significant interest. These AI systems must first build an internal representation of an environment, and then use it to simulate future events within that environment \cite{RunwayML24_GeneralWorldModels}, either in pixel representations \cite{Hu23arxiv_GAIA,Brooks24_Sora,Valevski24arxiv_GameNGen} or abstract latent future simulations \cite{Ha18neurips_RecurrentWorldModels,Hafner20iclr_Dreamer,Hafner21iclr_DreamerV2}. The most challenging setting is an \emph{interactive} world model, where the world is rolled out autoregressively \cite{Valevski24arxiv_GameNGen, Wu24neurips_SMART}, rather than in a single shot \cite{Brooks24_Sora, Polyak24arxiv_MovieGen}.

\boldparagraph{Diffusion-Based Traffic Simulation} SceneDiffuser \cite{Jiang24neurips_scenediffuser} demonstrated that a unified model could be used for both scene initialization and closed-loop rollout, and that amortized diffusion \cite{Zhang23arxiv_TEDi} could make diffusion more efficient and realistic. However, SceneDiffuser suffers from 3 key limitations. First, it predicts only agent features, rather than other environment features; second, it assumes known agent validity from logged data, limiting simulation duration to the length of logged data in WOMD; third, both the AV and world agents are jointly produced by a single model, potentially introducing collusion. In our work, we address all these limitations. Other methods also use diffusion for open-loop agent simulation \cite{Jiang23cvpr_MotionDiffuser, choi2023dice,Yang24arxiv_WorldCentricDiffusionTransformer, Guo23arxiv_SceneDM, Zhong23corl_CTG++,Niedoba23neurips_DiffJointIntNav}, closed-loop agent simulation \cite{Chang23arxiv_ControllableClosedLoopDiff, Huang24arxiv_VBD}, or for initial condition generation \cite{Lu24icra_SceneControl,Pronovost23neurips_ScenarioDiffusion,Pronovost23icraw_GenDrivingScenesDiffusion, Chitta24arxiv_SLEDGE, Sun24ral_DriveSceneGen}.

\boldparagraph{Agent Insertion and Deletion for Simulation} 
Most works assume a fixed set of agents throughout an entire scene \cite{Bergamini21icra_SimNet,Suo21cvpr_TrafficSim, Ngiam22icml_SceneTransformer, Niedoba23neurips_DiffJointIntNav, Mahjourian24icra_UniGen, Philion24iclr_Trajeglish, Jiang24neurips_scenediffuser,Huang24arxiv_VBD, Zhang24eccv_LearningSelfPlay, Sun24ral_DriveSceneGen, Chang2024eccv_SafeSim,Peng24eccv_AgentRLFineTuning, tan2024promptable}, reducing simulation to a two-stage process: scene initialization and subsequent rollout. Some of these works insert all agents before the start of the simulation, i.e. for a single fixed timestep, either all at once \cite{Pronovost23neurips_ScenarioDiffusion, Tan23corl_LCTGen, Sun24ral_DriveSceneGen}, or sequentially using conditional GANs \cite{Bergamini21icra_SimNet}, ConvLSTMs \cite{Tan21cvpr_SceneGen}, GMMs \cite{Feng22arxiv_TrafficGen}, or predicted occupancy grids \cite{Mahjourian24icra_UniGen}. This two-stage factorization approach has two limitations: First, that it causes simulation realism to degrade due to a lack of actor density, and second, that it cannot capture the real-world complexity of evolving and highly dynamic scenes. To our knowledge, no works have studied agent deletion using learned models.


To achieve long-duration simulation, CARLA \cite{Dosovitskiy17corl_CARLA}, SUMO \cite{Lopez18iits_SUMO}, and MetaDrive \cite{Li22tpami_Metadrive} rely on heuristics to insert and delete agents into the scene. For example, MetaDrive \cite{Li22tpami_Metadrive} procedurally generates maps and spawn points, assigns traffic vehicles to random spawn points on the map, and then recycles them if they stray too far from the AV. However, again, these simplistic heuristics cannot fully capture the complexity and diversity of real world traffic scenes. 

\begin{figure*}[!h]
\vspace{-3mm}
  \centering
    \includegraphics[height=10em,width=\linewidth]{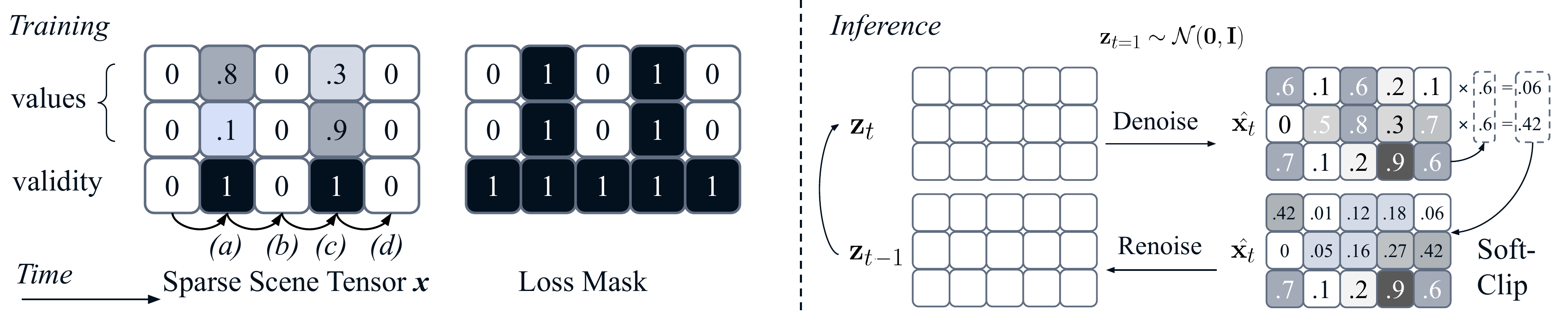}
    \caption{Learning \emph{sparse signals} with diffusion models. We illustrate the scene tensor using one single slice (for a single agent). Learning the validity field in this sparse tensor allows us to model (a) agent spawning, (b) occlusion, (c) disocclusion and (d) removal. \textbf{Left}: During training, we impute the corresponding values for all invalid steps to always be zero, before adding noise to train the denoiser. A corresponding loss mask is applied on the diffusion loss. \textbf{Right}: During inference, we adopt soft-clipping to multiply the intermediate denoised values by the predicted denoised validity, effectively interpolating with an all zero values vector, weighted by validity confidence.}
    \vspace{-3mm}
    \label{fig:learning-sparse}
\end{figure*}


\boldparagraph{Environment Simulation} Most methods for autonomous driving simulate only agent behavior and attributes, and not the surrounding environment. The dynamic environment itself can influence world agent and AV behavior, from traffic signals, to weather and time of day, to road hazards, debris, and construction. While data-driven sensor-simulation aims to generate a rendering of the environment \cite{Tancik22cvpr_BlockNeRF, Yang23cvpr_Unisim, Zhou24cvpr_DrivingGaussian}, we focus instead on a semantic, mid-level representation of the environment, such as traffic light states. 
To our knowledge, we are the first data-driven method to jointly simulate traffic light signal states and positions, as previous traffic signal control methods assume known signal positions~\cite{He24arxiv_TrafficSignalControl,Wei19arxiv_TrafficSignalControlSurvey,Liang19tvt_DeepRLTrafficLightControl}.
CARLA \cite{Dosovitskiy17corl_CARLA} controls traffic signals using heuristics. %

\section{Method}

\boldparagraph{Scene Tensor} We denote the scene tensor as $\bm{x}_i\in\mathbb{R}^{E_i\times \Tau\times D_i}$, where $E_i$ is the number of elements in the i-th scene tensor (e.g. agents, or traffic lights / signals) jointly modeled in the scene, $\Tau$ is the total number of modeled physical timesteps, and $D$ is the dimensionality of all the jointly modeled features. We learn to predict attributes for each element: for agents, these are validity $v$, positional coordinates $x, y, z$, heading $\gamma$, bounding box size $l, h, w$, and object type $k\in \{\text{AV, car, pedestrian, cyclist}\}$. For traffic lights, these are validity $v$, positional coordinates $x, y, z$ and a categorical traffic light state $s$. 
All features are normalized to $(-1, 1)$ range while agent types are one-hot encoded. All positional coordinates are normalized by the AV's ego pose. We frame all the tasks considered in \ours{} as multi-task inpainting tasks on these scene tensors, conditioned on an inpainting mask $\bar{\bm{m}_i}\in\mathbb{B}^{E_i\times \Tau\times D_i}$, the corresponding inpainting context values $\bar{\bm{x}_i}:=\bar{\bm{m}_i}\odot\bm{x}_i$, and a set of global contexts $\bm{c}$ (such as roadgraph).

\boldparagraph{Multi-Tensor} We define a multi-tensor $\mathcal{X}:=\{\bm{x}_i\}, \forall i$ as a collection of scene tensors. See Fig.~\ref{fig:multitensor} for an illustration of the multi-tensor structure and scene tensors. Without loss of generality, we learn $\mathcal{X}=\{\bm{x}_{\text{agent}}, \bm{x}_{\text{light}}\}$ for the joint distribution of agents and traffic lights.
We train a diffusion model to learn the conditional probability $p(\mathcal{X} | \mathcal{C})$ where $\mathcal{C}:=\{\bar{\bm{m}_i}, \bar{\bm{x}_i}, \bm{c}_i\}, \forall i$.
Note that we thus predict a validity mask $\bar{\bm{v}_i}\in{\mathbb{B}^{E_i, \Tau}}$ for a given element (agent or traffic signal) at a given timestep  (to account for there being $<E_i$ agents or lights in the scene or for occlusion).

\boldparagraph{Diffusion Preliminaries}
We adopt the notation and setup for diffusion models from \cite{Hoogeboom2023_SimpleDiffusion, Jiang24neurips_scenediffuser}. Below we denote all scene tensors as $\bm{x}$ and multi-tensor diffusion is a drop-in replacement of it with $\mathcal{X}$. Forward diffusion gradually adds Gaussian noise to $\bm{x}$. The noisy scene tensor at diffusion step $t$ can be expressed as $\mathbf{q}(\bm{z}_t|\bm{x}) = \mathcal{N}{(\bm{z}_t|\alpha_t \bm{x}, \sigma_t^2 \bm{I})}$,
where $\alpha_t$ and $\sigma_t$ are parameters controlling the magnitude and variances of the noise schedule under a variance-preserving model. Therefore $\bm{z}_t=\alpha_t \bm{x} + \sigma_t \bm{\epsilon}_t$, where $\bm{\epsilon}_t\sim\mathcal{N}(0, \bm{I})$. We apply the $\alpha$-cosine schedule where $\alpha_t=\cos (\pi t/2)$ and $\sigma_t = \sin (\pi t/2)$. At peak noise $t=1$, the forward diffusion process completely destroys the initial scene tensor $\bm{x}$ resulting in $\bm{z}_t = \bm{\epsilon}_t \sim \mathcal{N}{(0,\bm{I})}$. Assuming a Markovian transition process, the transition distribution is $q(\bm{z}_t|\bm{z}_s) = \mathcal{N}(\bm{z}_t|\alpha_{ts}\bm{z}_s,\sigma_{ts}^2\bm{I})$, where $\alpha_{ts}=\alpha_t/\alpha_s$ and $\sigma_{ts}^2=\sigma_t^2-\alpha_{ts}^2\sigma_s^2$ and $t>s$. The denoising process, conditioned on a single datapoint $\bm{x}$, can be written as
\begin{align}
    q(\bm{z}_s|\bm{z}_t,\bm{x}) = \mathcal{N}(\bm{z}_t|\bm{\mu}_{t\rightarrow s},\sigma_{t\rightarrow s}^2\bm{I}) ,
    \label{eqn:denosing_step}
\end{align}
where $\bm{\mu}_{t\rightarrow s}=\frac{\alpha_{ts}\sigma_s^2}{\sigma_t^2}\bm{z}_t + \frac{\alpha_s \sigma_{ts}^2}{\sigma_t^2}\bm{x}$ and $\sigma_{t\rightarrow s}=\frac{\sigma_{ts}^2\sigma_s^2}{\sigma_t^2}$. $\bm{x}$ is approximated using a learned denoiser $\hat{\bm{x}}$. Following \cite{Hoogeboom2023_SimpleDiffusion, Salimans2022_ProgressiveDistillation, Jiang24neurips_scenediffuser}, we adopt the \textit{v prediction} formulation, defined as $\bm{v}_t(\bm{\epsilon}_t, \bm{x})=\alpha_t \bm{\epsilon}_t - \sigma_{t}\bm{x}$. A model parameterized by $\bm{\theta}$ is trained to predict $\bm{v}_t$ from $\bm{z}_t$, $t$ and context $\mathcal{C}$: $\hat{\bm{v}}_t := \hat{\bm{v}}_{\theta}(\bm{z}_t, t, \mathcal{C})$. We can recover the predicted $\hat{\bm{x}}_t$ via $\hat{\bm{x}}_t = \alpha_t \bm{z}_t - \sigma_t \hat{\bm{v}}_t$. The entire model is end-to-end trained with a single loss function:
\begin{align}
    \mathbb{E}_{\substack{(\bm{x}, \mathcal{C})\sim \mathcal{D},  t\sim\mathcal{U}(0, 1), \\\bm{m}\sim \mathcal{M}, \epsilon_t\sim \mathcal{N}(0, \bm{I})}} [||(\hat{\bm{v}}_{\theta}(\bm{z}_t, t, \mathcal{C}) - \bm{v}_t(\bm{\epsilon}_t, \bm{x})\big)\cdot\bm{w}||_2^2] ,
    \label{eqn:loss_fn}
\end{align}
$\mathcal{D} = \{(\mathcal{X}, \mathcal{C})_j \mid j = 1, 2, \cdots, |\mathcal{D}|\}$ is the dataset containing paired agents and scene context data, $t$ is probabilistically sampled from a uniform distribution, and $\bm{w}\in \mathbb{B}^{E\times \Tau \times D}$ is a loss weighting term which we describe in more detail below. $\mathcal{M} = \{\bar{\bm{m}}_{\text{bp}} \odot \bar{\bm{m}}_{\text{control}}, \bar{\bm{m}}_{\text{scenegen}} \odot \bar{\bm{m}}_{\text{control}}\}$ is the set of inpainting masks for the varied tasks elaborated below.

\label{sec:tasks}
\boldparagraph{Tasks} We formulate the various tasks, such as scene generation (SceneGen) and behavior prediction (BP) as different inpainting tasks. Following SceneDiffuser \cite{Jiang24neurips_scenediffuser}, the BP inpainting mask $\bm{m}_{\text{bp}}$ has 1 for all history steps and 0 for all future steps. SceneGen mask $\bm{m}_{\text{scenegen}}$ consists of 1 for randomly chosen context agents and 0 for agents to predict. The control mask $\bm{m}_{\text{control}}$ which consists of randomly sampled $\{0, 1\}$, is applied on top of either the SceneGen or BP task for further controllability. This work is a special case of BP with additional agent validity prediction.

\boldparagraph{Architecture} While different event types may differ in the number of entities to predict and their feature dimensions, we adapt the same context encoder and Transformer denoiser backbone architecture as SceneDiffuser \cite{Jiang24neurips_scenediffuser} by homogenizing different scene tensors. We first project different scene tensors to the same hidden dimension, followed by concatenating along the `elements' axis (see Fig. \ref{fig:multitensor}). After adopting the SceneDiffuser \cite{Jiang24neurips_scenediffuser} backbone, we apply the reverse process to split and unproject them into the respective scene tensors.


\boldparagraph{Learning Sparse Tensors} One of the major technical contributions of this work is a method for predicting sparse tensors using diffusion. While predicting sparse tensors is of critical importance in this work for learning agent spawning, removal and occlusion, the problem is generic pertaining to learning sparse signals using diffusion models.

Given a sparse tensor $\bm{x}\in\mathbb{R}^{E\times \Tau \times D}$, we decompose it into the values: $\mathcal{V}(\bm{x})=\bm{x}[..., :-1]\in\mathbb{R}^{E\times \Tau \times (D-1)}$ and the validity mask: $\mathcal{M}(\bm{x})=\texttt{Clip}(\bm{x}[..., -1], -1, 1)/2 + 0.5 \in \mathbb{B}^{E\times \Tau}$. The affine transformation is to reverse the initial normalization. We define its inverse to be $\mathcal{M}^{-1}(\bm{x}) = 2\bm{x}-1$. When the validity is \texttt{False}, the corresponding value are arbitrary, whereas when validity is $\texttt{True}$, the corresponding values are meaningful. We seek to jointly predict the values and validity mask. This presents a set of challenges to conventional diffusion model training. How do we supervise the training of values if some corresponding values do not have ground truth (since they are invalid)? Two alternatives arise: (1) impute all invalid values to be zero, and train as if it's a dense tensor, or (2) leave the invalid bits in the values unsupervised. We find that neither of these two approaches can work. Imputation of zeros for invalid values creates significant discontinuities in the signal, leading to unstable model training. Alternatively, if one leaves the invalid bits unsupervised, they are recurrently fed into the denoiser at inference time, leading to very rapid slippage into out-of-distribution values.

We implement a simple yet effective alternative to these approaches, as described in Fig. \ref{fig:learning-sparse}. We first cast all values corresponding to invalid steps to zero with $\bm{x} \leftarrow \bm{x} \cdot \mathcal{M}(\bm{x})$. Then we compute the loss based on Eqn. \ref{eqn:loss_fn}. For weight $\bm{w}$, we apply the loss mask in as illustrated in Fig. \ref{fig:learning-sparse}, where all features in valid steps are supervised, and only the validity feature in invalid steps are supervised. During inference, we first sample $\bm{z}_{t=1}\sim\mathcal{N}(0, \bm{I})$. Then at each denoising step, we first predict the denoised solution at that step: $\hat{\bm{x}_t}=\alpha_t\bm{z}_t - \sigma_t \hat{\bm{v}_{\theta}}(\bm{z}_t, t, \mathcal{C})$. Then we apply a step we coin \emph{soft clipping} which we describe below. Finally, we renoise the result to a lower noise level $t-1$ with $\bm{z}_{t-1}\sim \mathcal{N}(\bm{z}_{t-1}|\alpha_{t-1}\hat{\bm{x}_{t}}, \sigma^2_{t-1}\bm{I})$. 

We find the \textit{clipping} step to be the most crucial inference-time trick for generating sparse tensors. We contrast soft clipping with several variants:
\begin{itemize}
    \item soft clipping: $\hat{\bm{x}_t}\leftarrow \texttt{Concat}(\mathcal{V}(\hat{\bm{x}_t}) * \mathcal{M}(\hat{\bm{x}_t}), \mathcal{M}(\hat{\bm{x}_t}))$
    \item no clipping: $\hat{\bm{x}_t}\leftarrow\hat{\bm{x}_t}$\\
    Given the two clipping functions for mask and value:\\
    \scalebox{1}[1.0]{$\bm{m}_{clipped} \leftarrow \mathcal{M}^{-1}(\texttt{Where}(\mathcal{M}(\hat{\bm{x}}_t) < 0.5, \bm{0}, \bm{1}))$}\\
    \scalebox{1}[1.0]{$\bm{v}_{clipped} ~~\leftarrow \texttt{Where}(\mathcal{M}(\hat{\bm{x}}_t) < 0.5, \bm{0}, \mathcal{V}(\hat{\bm{x}_t}))$}
    \item hard clipping: $\hat{\bm{x}_t}\leftarrow\texttt{Concat}(\mathcal{V}(\hat{\bm{x}_t}), \bm{m}_{clipped})$
    \item hard-validity clip.: $\hat{\bm{x}_t}\leftarrow\texttt{Concat}(\bm{v}_{clipped}, \bm{m}_{clipped})$
    
\end{itemize}
We show in Sec. \ref{subsec:ablation} that soft clipping is the most effective strategy that allows stable training and inference with minimal additional changes.

\section{Experiment}

\subsection{Trip-level Traffic Simulation Setup}

\boldparagraph{Dataset} We use the Waymo Open Motion Dataset (WOMD)\cite{Ettinger21iccv_WOMD} for our trip-level traffic simulation experiments. WOMD includes tracks of all agents and corresponding vectorized maps in each scenario, and it offers a large quantity of high-fidelity object behaviors and shapes produced by a state-of-the-art offboard perception system.
Each scenario in WOMD consists of 91 timesteps with a frequency of 10Hz, leading to a 9.1 second scenario.
Although the scenario clips are much shorter than our trip-level simulation route, they contain all the critical agent behaviors (driving, entering, exiting, occlusion) and traffic light states (position detection, lane association, state changes). 
This feature allows us to use these short clips to train \ours{} models that can simulate trip-level scenarios much longer than 9.1 seconds.
%
However, during trip-level rollouts ($> 9.1 s$), we note that agents easily run out of the map and roadgraph extent, as the original WOMD dataset only contains map regions that cover where the AV can reach in 9.1 seconds.
To conduct trip-level simulation, we asked for expanded maps from the WOMD dataset creators (all map elements within circles of 1km radius around any portion of the AV’s trajectory) to generate a map-extended dataset that we call \emph{WOMD-XLMap}.

%

\boldparagraph{World Model vs. Planner} Real simulation use cases require interaction between two disjoint models -- a planner and a simulator (world model) \cite{Montali23neurips_wosac,Peng24eccv_AgentRLFineTuning, Karnchanachari24icra_nuPlan}. 
Specifically, the planner controls the AV's movement given the environment and other agents' movement. On the other hand, the world model controls the traffic lights and all other background agents' movement given the AV's movement.
At each rollout step, the planner can only observe the world model's history output and cannot obtain its future predictions, and
vice versa for world model.
In other words, the planner and world model observe each other's predictions only after we rollout their predictions in the environment.
In the case where we use the same method as both planner and world model, we ensure they do not share the same predictions by setting different seeds for random sampling. 


\boldparagraph{Method Comparisons} 
We first compare with the \emph{SceneDiffuser}\cite{Jiang24neurips_scenediffuser} model.
Unlike \ours{}, SceneDiffuser does not model agent validity nor traffic light features.
Therefore, during prolonged rollouts with SceneDiffuser, we simply assume that all the agents valid at the current step will remain valid in the future, while setting any future traffic light features as invalid.  
We also compare with the Intelligent Driver Model (\emph{IDM}) \cite{Treiber00pre_IDM, Gulino23neurips_Waymax} model.
To set the routes for IDM to drive for each agent, we start with each agent's initial location and randomly select a valid path with the lane graph on the map.
For validity, we set all the agents' future validity to remain the same as their current validity.
In our main experiment, we test each possible combination of planner and world model using the three methods.

\begin{table*}[!h]
\vspace{-2mm}
\caption{Simulation realism under long rollouts (60s). Numbers are JS-divergence between simulated and logged distributions ($\downarrow$). Composite is the average of all metrics except TL Violation and TL Transition.}
\label{main_exp_table}
\centering
\begin{adjustbox}{width=\linewidth}
\begingroup
\begin{tabular}{ll ccccccccccc}
\toprule
\rowcolorize  &  & \# Valid  & \# Entering  & \# Exiting  & Entering  & Exiting   & Offroad  &  Collision  & Average  & TL  & TL  & \\
\rowcolorize World Model & Planner &  Agents &  Agents &  Agents &  Distance &  Distance  &  Rate &   Rate &  Speed &  Violation & Transition & Composite\\

\midrule
IDM & IDM  & 0.4028 & 0.6357 & 0.5125 & 0.3780 & 0.5253 & 0.3578 & \textbf{0.3652} & 0.6570 & - & - & 0.4793 \\
\rowcolorize SceneDiffuser & IDM & 0.5701 & 0.7027 & 0.5767 & 0.3830 & 0.3296 & 0.2765 & 0.3778 & 0.6213 & - & - & 0.4797 \\
SceneDiffuser++ & IDM & \textbf{0.3132} & \textbf{0.1947} & \textbf{0.2059} & \textbf{0.1620} & \textbf{0.1549} & \textbf{0.2428} & 0.4361 & \textbf{0.5908} & 0.1582 & 0.0589 & \textbf{0.2878} \\
\hline
\rowcolorize IDM & SceneDiffuser & 0.2941 & 0.7331 & 0.7279 & - & - & \textbf{0.0846} & \textbf{0.1017} & 0.4917 & - & - & - \\
SceneDiffuser & SceneDiffuser & 0.4532 & 0.7114 & 0.6275 & - & 0.2759 & 0.2056 & 0.3217 & 0.4036 & - & - & - \\
\rowcolorize SceneDiffuser++ & SceneDiffuser & \textbf{0.2206} & \textbf{0.1409} & \textbf{0.1526} & \textbf{0.1668} & \textbf{0.1494} & 0.1345 & 0.4940 & \textbf{0.3858} & 0.1596 & 0.0264 & \textbf{0.2306} \\
\hline
IDM & SceneDiffuser++ & 0.3967 & 0.6255 & 0.5170 & 0.5250 & 0.5384 & 0.2056 & \textbf{0.2990} & \textbf{0.2840} & - & - & 0.4234 \\
\rowcolorize SceneDiffuser & SceneDiffuser++ & 0.5373 & 0.6921 & 0.5718 & 0.3746 & 0.2514 & 0.2384 & 0.3812 & 0.4562 & - & - & 0.4389 \\
SceneDiffuser++ & SceneDiffuser++ & \textbf{0.3053} & \textbf{0.2120} & \textbf{0.2085} & \textbf{0.1183} & \textbf{0.1094} & \textbf{0.1595} & 0.4194 & 0.3061 & 0.1625 & 0.0448 & \textbf{0.2423} \\
\bottomrule
\end{tabular}
\endgroup
\end{adjustbox}
\end{table*}

\boldparagraph{Metrics}
For long rollouts, we end up with significant divergence between the logged scene and the propagated scene rollout. Accordingly, it does not make sense to constrain the simulation to adhere closely to the logged data, as done in WOSAC \cite{Montali23neurips_wosac}. We also have no 1:1 correspondence between agents, as agents may enter and exit the scene freely.

We use a sliding evaluation window over temporal segments of our long-duration rollouts and ensure that each window has the same temporal length as the log scenario.
Then, at each temporal window, we collect the simulated metric value (e.g., number of valid agents) from all simulated scenarios to a list of sim metrics. We also collect all the metric values for the log data to a list of log metrics. 
We fit two histograms to the sim and log metric values. To measure the realism of the sim features, we compute the Jensen–Shannon (JS) Divergence \cite{Lin91tit_DivergenceShannonEntropy} between these histograms. 
Lower divergence between histograms indicates more realistic simulated scenarios. Then we compute the mean value over all the divergence values for all windows.

Here we introduce the features over which we compute distributional metrics in our experiments: 1) \emph{\# Valid Agents}: the number of agents that have at least one timestep that is valid in the scenario window; 2) \emph{\# Entering/Exiting Agents}: the number of agents that are inserted or removed during the scenario window, respectively; 3) \emph{\# Entering/Exiting Distance}: the distance to the AV of the entering or exiting agents at the first or last valid timestep in the scenario, respectively; 4) \emph{Offroad Rate}: the fraction of all valid agents located offroad (e.g., in parking lots); 5) \emph{Collision Rate}: the fraction of all valid agents that ever collide with other agents; 6) \emph{Average Speed}: the average speed for all the valid agents in the scenario window; 7) \emph{TL Violation}: the fraction of all valid agents that violate traffic light rules; 8) \emph{TL Transition}\footnote{
For \textit{TL Transition}, we directly compute the divergence between log and sim transition probability matrices computed over all scenarios.}: the transition probability between different traffic light states (e.g., from red to green). 
Finally, we compute a \textit{Composite} score that is the average of all the metrics.

\boldparagraph{Simulation Configuration} We follow the ``Full AR'' inference scheme of SceneDiffuser \cite{Jiang24neurips_scenediffuser}. We vary two key simulation parameters: 1) \# rollout steps: the total number of timesteps to rollout, and 2) \# replan steps: the number of timesteps between each planner / world model replanning. The smaller the \# replan steps, the more frequently the planner and world model are executed and interact with each other.
In our main experiments, we set \# rollout steps = 600 (60 seconds @ 10Hz) and \# replan steps = 40, but we also explore the effects of replan frequencies in Sec. \ref{subsec:ablation}. Please refer to the Appendix for training and model details.

\begin{table*}[!h]
\vspace{-3mm}
\caption{Controlled evaluation of simulation configurations. \ours{} serves as both planner and world model.}
\label{sim_cfg_table}
\centering
\begin{adjustbox}{width=\linewidth}
\begingroup
\begin{tabular}{cc ccccccccccc}
\toprule
\rowcolorize \# Rollout  & \# Replan  & \# Valid  & \# Entering  & \# Exiting  & Entering  & Exiting   & Offroad  &  Collision  & Average  & TL   & TL  & \\
\rowcolorize  Steps &  Steps & Agents &  Agents &  Agents &  Distance & Distance  &  Rate &   Rate &  Speed &  Violation  & Transition & Composite\\
\midrule
600 & 10  & 0.3463 & 0.2308 & 0.2199 & 0.0952 & 0.1231 & 0.1581 & \textbf{0.3118} & \textbf{0.2681} & \textbf{0.1526} & 0.0584 & \textbf{0.1867} \\
\rowcolorize 600 & 20  & 0.3286 & 0.2211 & 0.2165 & \textbf{0.0910} & \textbf{0.1030} & 0.1729 & 0.3702 & 0.2872 & 0.1539 & \textbf{0.0448} & 0.1937 \\
600 & 80  & \textbf{0.2775} & \textbf{0.1853} & \textbf{0.1840} & 0.1522 & 0.1356 & \textbf{0.1461} & 0.4478 & 0.3204 & 0.1845 & 0.0668 & 0.2102 \\
\hline
\rowcolorize 300 & 40  & \textbf{0.2526} & \textbf{0.1947} & \textbf{0.1860} & \textbf{0.1195} & \textbf{0.1075} & \textbf{0.1165} & 0.4128 & \textbf{0.2687} & \textbf{0.1579} & 0.0396 & \textbf{0.1936} \\
1200 & 40  & 0.3457 & 0.2268 & 0.2259 & 0.1195 & 0.1129 & 0.1950 & 0.4172 & 0.3284 & 0.1715 & \textbf{0.0312} & 0.2213 \\
\rowcolorize 3000 & 40  & 0.4018 & 0.2758 & 0.2736 & 0.1248 & 0.1235 & 0.2084 & \textbf{0.4104} & 0.2993 & 0.2243 & 0.0468 & 0.2468 \\
\bottomrule
\end{tabular}
\endgroup
\end{adjustbox}
\end{table*}

\vspace{-1em}


\begin{table*}[!h]
\caption{Controlled evaluation of SceneDiffuser++ inference time validity decoding strategies, as measured by JS Divergence ($\downarrow$).}
\label{decoding_study_table}
\centering
\begin{adjustbox}{width=\linewidth}
\begingroup
\begin{tabular}{c ccccccccccc}
\toprule
\rowcolorize  & \# Valid & \# Entering  & \# Exiting  & Entering & Exiting   & Offroad  &  Collision  & Average  & TL  & TL  &  \\
\rowcolorize Prediction Mode &  Agents &  Agents &  Agents &  Distance &  Distance  & Rate &   Rate & Speed & Violation & Transition & Composite \\

\midrule
Hard Clipping & 0.4927 & 0.4776 & 0.4094 & \textbf{0.1156} & 0.1245 & \textbf{0.0992} & \textbf{0.2602} & \textbf{0.2664} & 0.2099 & \textbf{0.0429} & 0.2498 \\
Hard-Validity Clipping & 0.5963 & 0.6510 & 0.5502 & 0.1741 & 0.1641 & 0.2072 & 0.2830 & 0.2780 & 0.2379 & 0.0435 & 0.3185 \\
No Clipping & \textbf{0.2426} & \textbf{0.2035} & 0.2139 & 0.1425 & 0.1029 & 0.3026 & 0.6697 & 0.3685 & 0.3123 & 0.1044 & 0.2663 \\
Soft Clipping & 0.3053 & 0.2120 & \textbf{0.2085} & 0.1183 & \textbf{0.1094} & 0.1595 & 0.4194 & 0.3061 & \textbf{0.1625} & 0.0448 & \textbf{0.2046} \\
\bottomrule
\end{tabular}
\endgroup
\end{adjustbox}
\end{table*}

\subsection{Main Results}
We show the main result of trip-level traffic simulation in Table~\ref{main_exp_table}.
We group different experiment settings by which planner is used, and compare the metric results for the rollouts using different world models. Note that some entries are not available in this table. 
Because SceneDiffuser and IDM do not insert agents into the scene after the first scenario window, their Entering Distance and Exiting Distance results are poor, as expected; accordingly, when using SceneDiffuser Planner, IDM and SceneDiffuser world models, we don't report their entering distance in Table~\ref{main_exp_table}. 

\begin{figure*}[!h]
  \centering
    \includegraphics[trim={0 0 0 0},clip,width=1.0\linewidth]{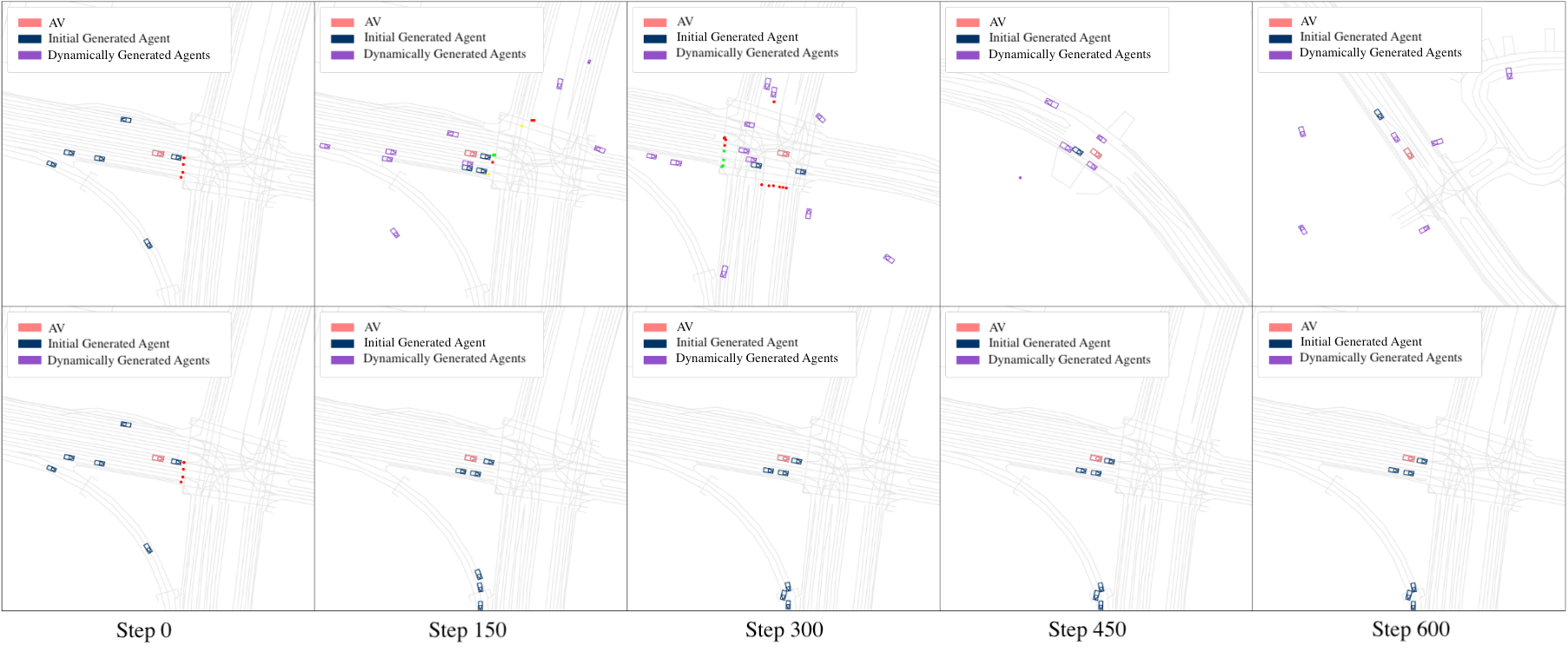}
    \caption{Visualization of a 60s rollout with \ours{} (top row) vs. SceneDiffuser, where all agents become stuck in a single location (bottom row). Note that SceneDiffuser++ predicts a variable number of traffic lights to be observable (visible) at any timestep; this is also observed in logged data at times.}
    \vspace{-1em}
    \label{fig:rollout2}
\end{figure*}

Our model achieves significantly better performance in all metrics that relate to agent insertion and removal.
For example, when we use IDM as the planner, using \ours{} as the world model leads to much more realistic distributions of the number of valid, entering and exiting agents, as well as entering and exiting agents' distances. These results indicate that \ours{} yields superior performance for predicting when and where to insert and remove agents.
In contrast, using IDM or SceneDiffuser models leads to much higher divergence between real and simulated distributions.
This result shows it is necessary to predict agent validity for trip-level simulation.

We observe that using our model as a world model leads to, in aggregate, better Average Speed likelihood of the scenario. 
This is mainly due to the fact that our model is able to predict realistic agent insertion and removal. When agents are able to dynamically appear and exit the scenario, we allow the model to focus more on realistic agent behaviors, e.g. their speed. 
On the other hand, when the model has to predict features for all the agents that appear in the current step in the future, it has to maintain all the agents' proximity to the AV. Consequently, for SceneDiffuser, we observe all the agents tend to become static during trip-level simulation. We show this effect in Figure~\ref{fig:rollout2}.

We also note that the Offroad Rate and Collision Rate of our model when used as a world model is worse than that of using IDM and SceneDiffuser. 
There are a few reasons for this performance:
First, during rollout, \ours{} will insert agents into the scenario, regardless of how the planner drives before the next replan step. 
Therefore, it is possible that \ours{} will insert agents onto the route of the planner in future steps. 
However, if no agents are inserted in the scenario (for IDM and SceneDiffuser), all the agents will follow their historic trajectories and drive on safe routes, meaning collisions are less likely.
We show in Table~\ref{sim_cfg_table} that with more frequent replanning, our model leads to a much better collision rate metric.
Additionally, we found that \ours{} tends to insert a large amount of agents in parking lots that stay parked. These generated parked agents lead to worse offroad metrics.

\subsection{Additional Analysis}\label{subsec:ablation}

\boldparagraph{{Inference-time Clipping}}
In Table~\ref{decoding_study_table}, we present an ablation on soft vs. hard vs. hard-validity clipping for generating sparse tensors.
We observe that only soft clipping of features leads to favorable distributions of the number of valid, entering and exiting agents.
This indicates that any hard clipping with the validity value on the feature will render the model unable to reflect the agent validity distribution.
In addition, we also show the results of a model trained to directly predict invalid agents' features to be 0, and not using clipping during inference (third row), but find this method leads to a higher collision rate, offroad rate, and TL violation rate, along with inferior TL transitions, demonstrating unstable feature prediction values.

\boldparagraph{{Simulation Configurations}}
In Table~\ref{sim_cfg_table}, we compare the results when using \ours{} as both world model and planner under different \# rollout steps and \# replan steps. 

We first show a comparison of different replan steps in the first three rows.
We observe that: 1) with more frequent replanning (smaller \# replan steps), our model achieves better Collision Rate and Average Speed. This is because when the world model and planner can interact more frequently, they are more reactive to each other's behavior.
2) with less frequent replanning (larger \# replan steps), our model leads to better agent insertion and removal behavior. 
With less frequent replanning, the world model has more timesteps to control into the future, which allows \ours{} to better plan when and where to insert agents over the full sequence. On the other hand, with a high replanning rate, only agents that will be predicted to enter into the scenario in the first few timesteps will be inserted, leading to an inferior distribution of agent validity.

In the next three rows, we show the ablation with the same replan frequency, but different planning horizons, from 30 seconds to 300 seconds.
We observe that overall, the realism metrics drop when the model is rolled out over longer horizons.
This is mainly due to error from the autoregressive rollout aggregating over time when rolling out over long horizons.
Note that although the realism of the number of entering and exiting agents degrades over time, the respective entering or exiting distances stay quite stable.
This might indicate that the aggregated error affects agent insertion timing more than insertion position.

\begin{figure}
    \subfloat[Inserted agent exits a parking lot.]{
        \includegraphics[width=0.22\textwidth]{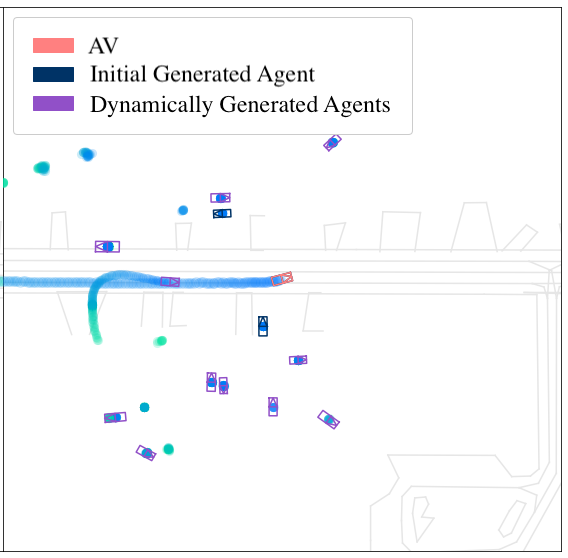}
    }
    \subfloat[Inserted agent far away from AV.]{
        \includegraphics[width=0.22\textwidth]{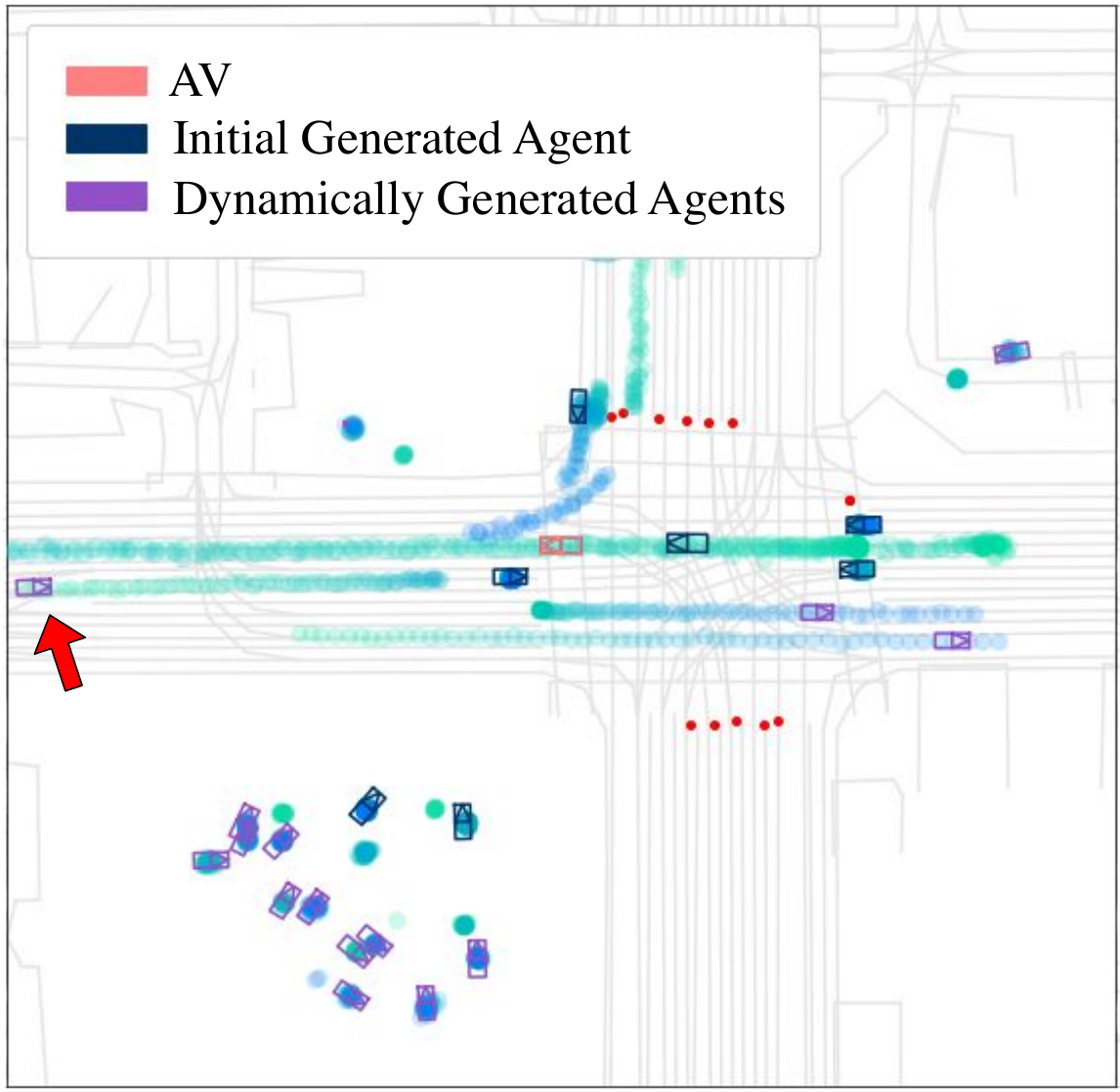}
    }

    \subfloat[Inserted agents stop at red light.]{
        \includegraphics[width=0.22\textwidth]{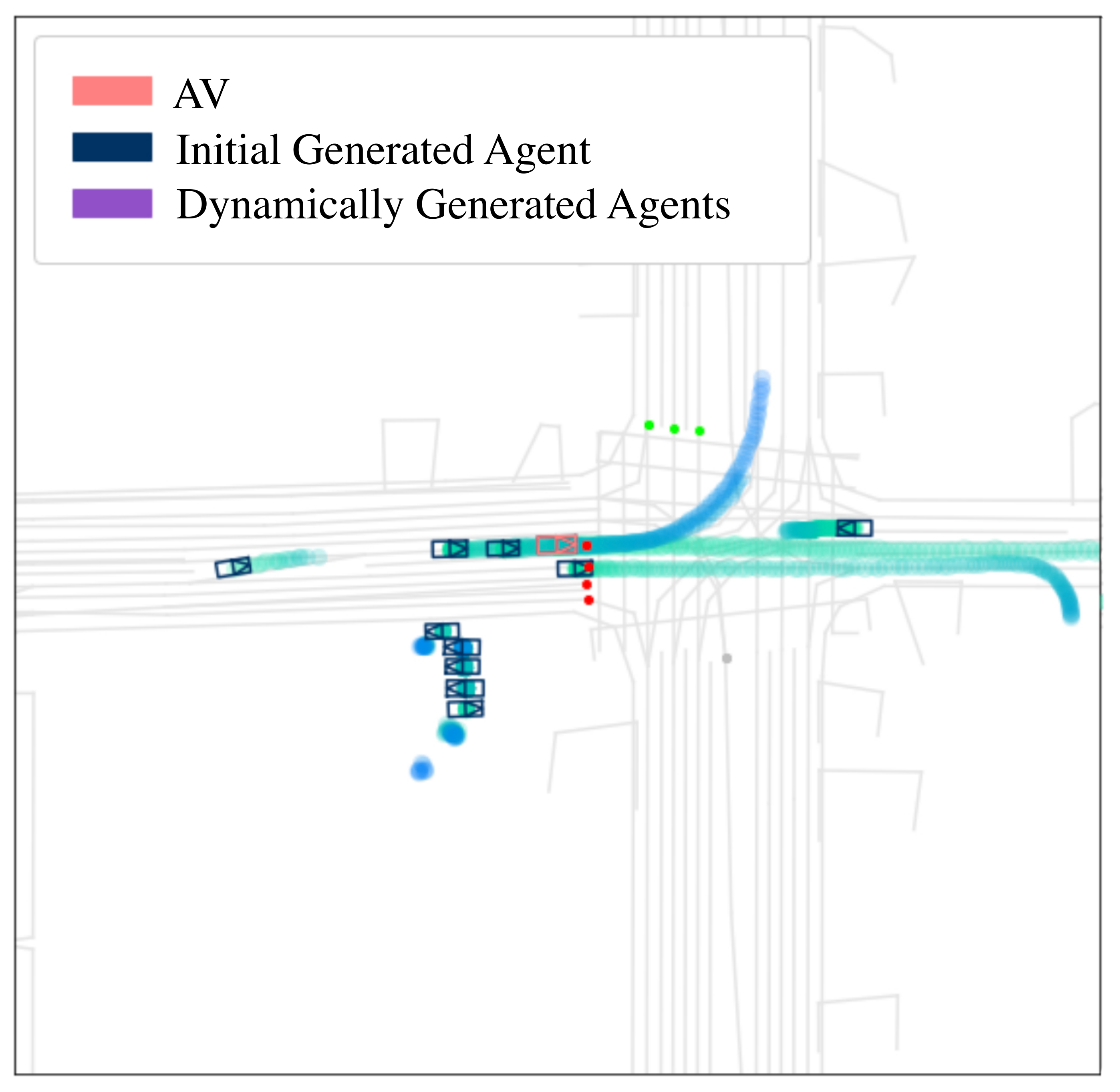}
    }
    \subfloat[Agents continue when light is green.]{
        \includegraphics[width=0.22\textwidth]{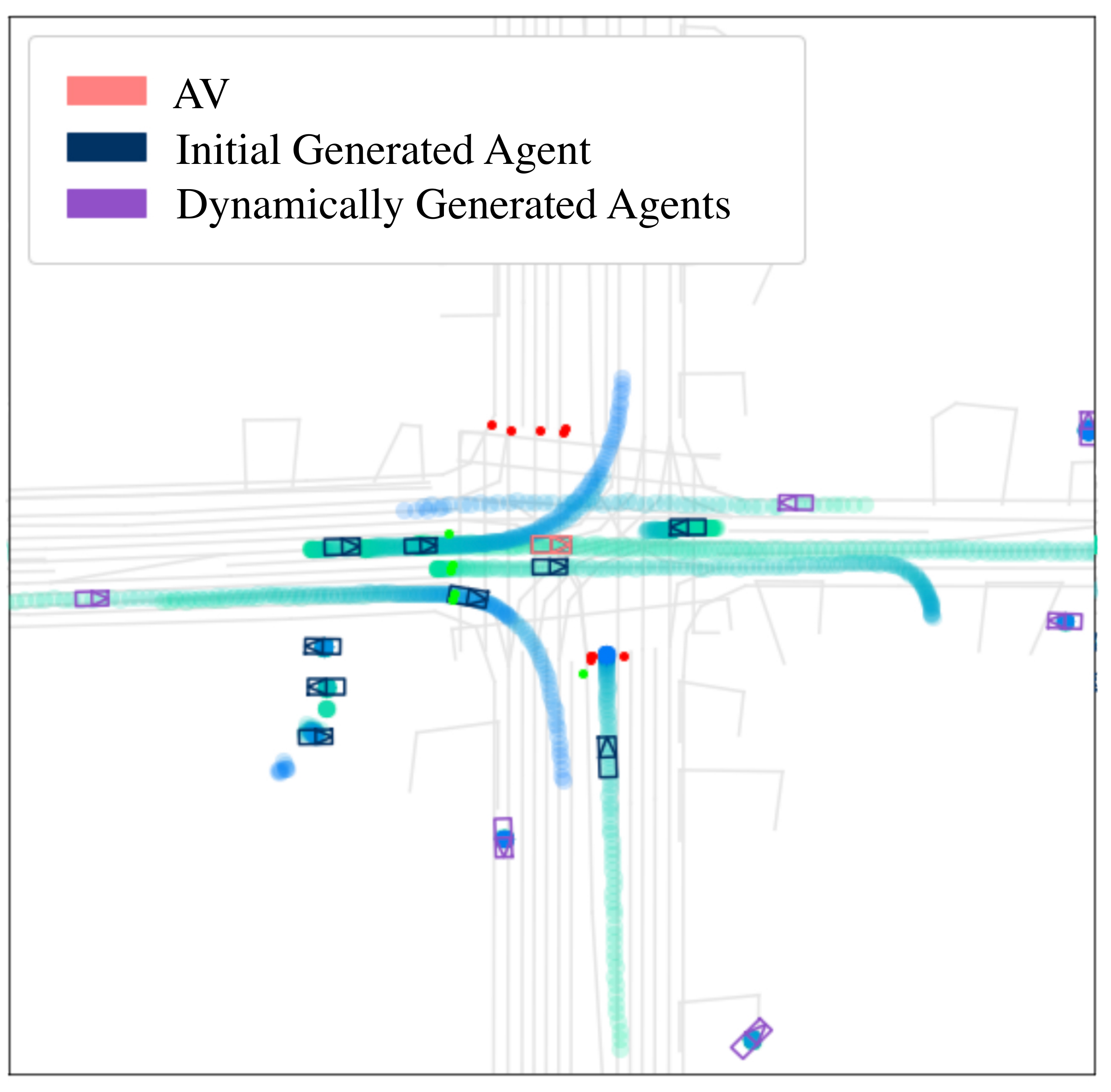}
    }
    \caption{Plots of \ours{}  results (\textcolor{waymogreen}{green}$\rightarrow$\textcolor{waymoblue}{blue} indicates temporal progression).}
    \label{fig:results}
\vspace{-3mm}
\end{figure}

\boldparagraph{{Metric Window Curves}}
In Fig. \ref{fig:curve}, we plot the values of two metrics in Table~\ref{main_exp_table} over simulation timesteps for all world model methods using SceneDiffuser++ as planner.
Our model achieves the best performance over all timesteps.

\boldparagraph{{Qualitative Results}}
In Fig. \ref{fig:rollout2} we show examples from 60-second rollouts of our model vs. SceneDiffuser, where we uniformly sample 5 frames from the total 600 steps of each rollouts. 
For both our model and SceneDiffuser, we roll out using the same model for both world model and planner.
It is obvious that our model achieves a realistic trip-level rollout across a large map area with dynamic traffic lights, while SceneDiffuser gets stuck in the starting location, as seen in Figure~\ref{fig:rollout2}.
This is mainly due to two reasons: 1) SceneDiffuser does not predict future traffic light location and states, leading to confusion of the AV in the intersection without any traffic lights. 2) SceneDiffuser does not model agents exiting the scenario, therefore it is forced to keep all the agents within the visible range of the AV. Note how the agents in the bottom of the scenario were forced to unrealistically stop in order to keep themselves in the scenario.
In comparison, \ours{} deals with these issues with a unified model and makes trip-level simulation possible.

\begin{figure}
    \centering    \includegraphics[width=0.9\linewidth]{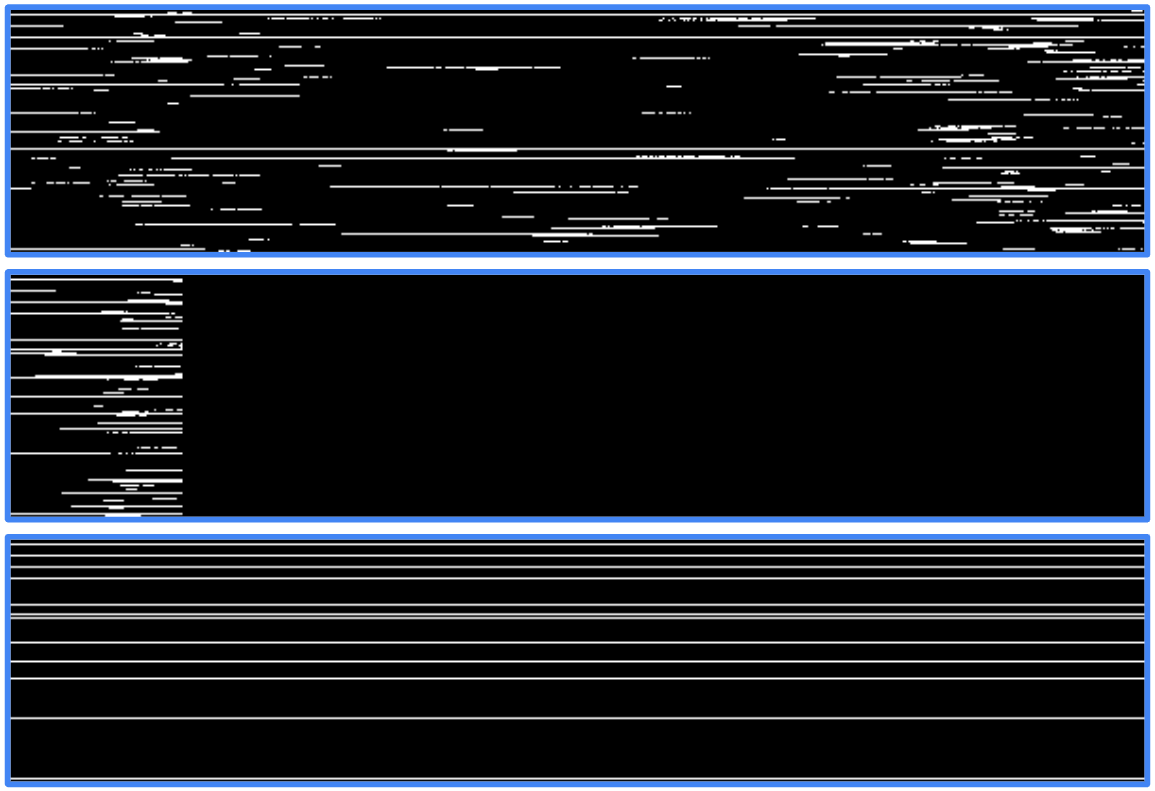}
    \caption{Plot of predicted validity for all 128 agents across 600 steps. X-axis is time, Y-axis is agent ID. White indicates that the agent is valid at that timestep. Top: SceneDiffuser++, Middle: Ground-truth Log (91 steps), Bottom: IDM, SceneDiffuser.}
    \label{fig:validity_plot}
    \vspace{-1em}
\end{figure}

\begin{figure}
    \centering
    \includegraphics[width=\linewidth]{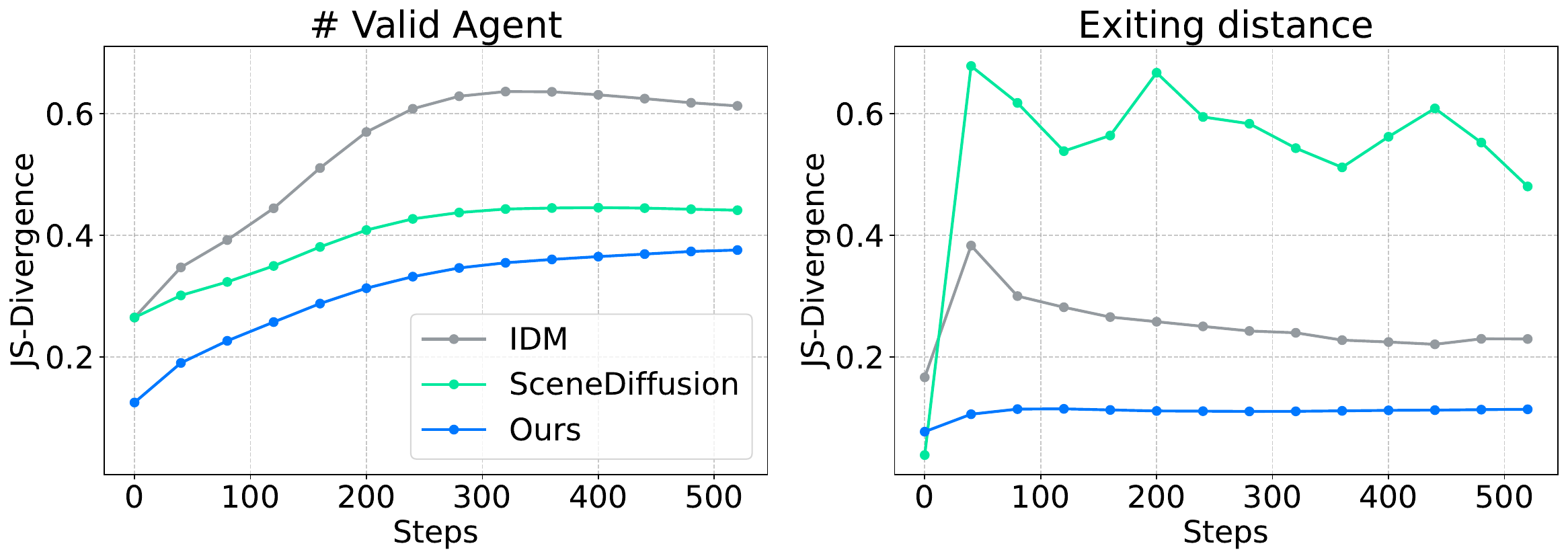}
    \caption{Divergence over simulation steps curve (lower is better). }
    \label{fig:curve}
    \vspace{-1em}
\end{figure}

\begin{figure}
    \centering
    \includegraphics[trim={45em 5em 65em 6em},clip,width=\linewidth]{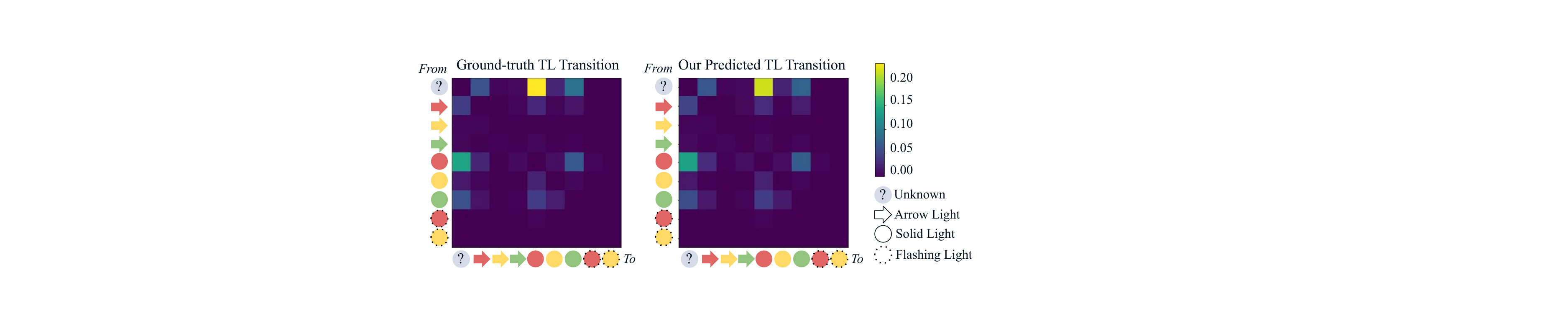}
    \caption{Traffic light transition probability matrix for log vs ours.}
    \label{fig:tl_transition}
    \vspace{-2em}
\end{figure}

Next, we display realistic generated agent behaviors. In Fig. \ref{fig:results}(a) we show that agents inserted into a parking lot can realistically navigate onto the main road and merge into traffic. Fig. \ref{fig:results}(c) and Fig. \ref{fig:results}(d) show that inserted agents comply with traffic light rules, indicating high realism of agent and traffic light interaction. Lastly, Fig. \ref{fig:results}(b) shows generated agents can be inserted far from the AV.

\boldparagraph{{Validity Prediction}}
We visualize the predicted agent validity through 600 steps in Fig. \ref{fig:validity_plot}, comparing our model's output, the logged `ground truth', and the pattern that IDM and SceneDiffuser both produce.
Our model is able to insert and remove agents with realistic validity patterns that are very close to the ground-truth in the first 91 steps. On the other hand, IDM and SceneDiffuser only follow the last-step history validity, leading to a quite unnatural validity pattern. 
Finally, note that our model is able to insert a new agent to an agent row that was previously occupied by a removed agent (e.g., the last few rows), as long as the previous agent was removed longer ago than SceneDiffuser++'s history horizon.
In this way, our method is able to insert any number of agents beyond the total number of agent indices by reusing any agent index where an agent was removed.

\boldparagraph{{Traffic Light Transition}}
We visualize the traffic light state transition probability matrix in Figure~\ref{fig:tl_transition}, where left is the logged ground-truth and right is \ours{} prediction.
In these figures, along the y-axis is the starting traffic light state and along the x-axis the ending traffic light state.
We observe that \ours{} traffic light state predictions rigorously follow the ground-truth state transition probability.
Note that we remove all the state self-transitions (the diagonal entries) for clearer visualization.

\vspace{-1em}

\section{Conclusion}
We have introduced \ours{}, a scene-level diffusion prior designed for city-scale traffic simulation.
\ours{} is a unified world model that enables trip-level long simulations with dynamic agent generation, occlusion reasoning, removal and traffic light simulation.
We demonstrate \ours{} has strong performance for long-term traffic simulation.
We hope our work leads to more realistic trip-level simulation to improve AV safety.



{
    \small
    \bibliographystyle{ieeenat_fullname}
    \bibliography{main}
}


\newpage
\appendix
\section{Appendix}

\subsection{Video}

We provide a \href{https://youtu.be/J70R3wxPQTc}{video}\footnote{\url{https://youtu.be/J70R3wxPQTc}} that features a brief and intuitive overview of:

\begin{figure*}[!h]
    \centering
    \newcommand{\margin}{0px} 
    \includegraphics[trim={\margin, \margin, \margin, \margin},clip,width=0.96\textwidth]{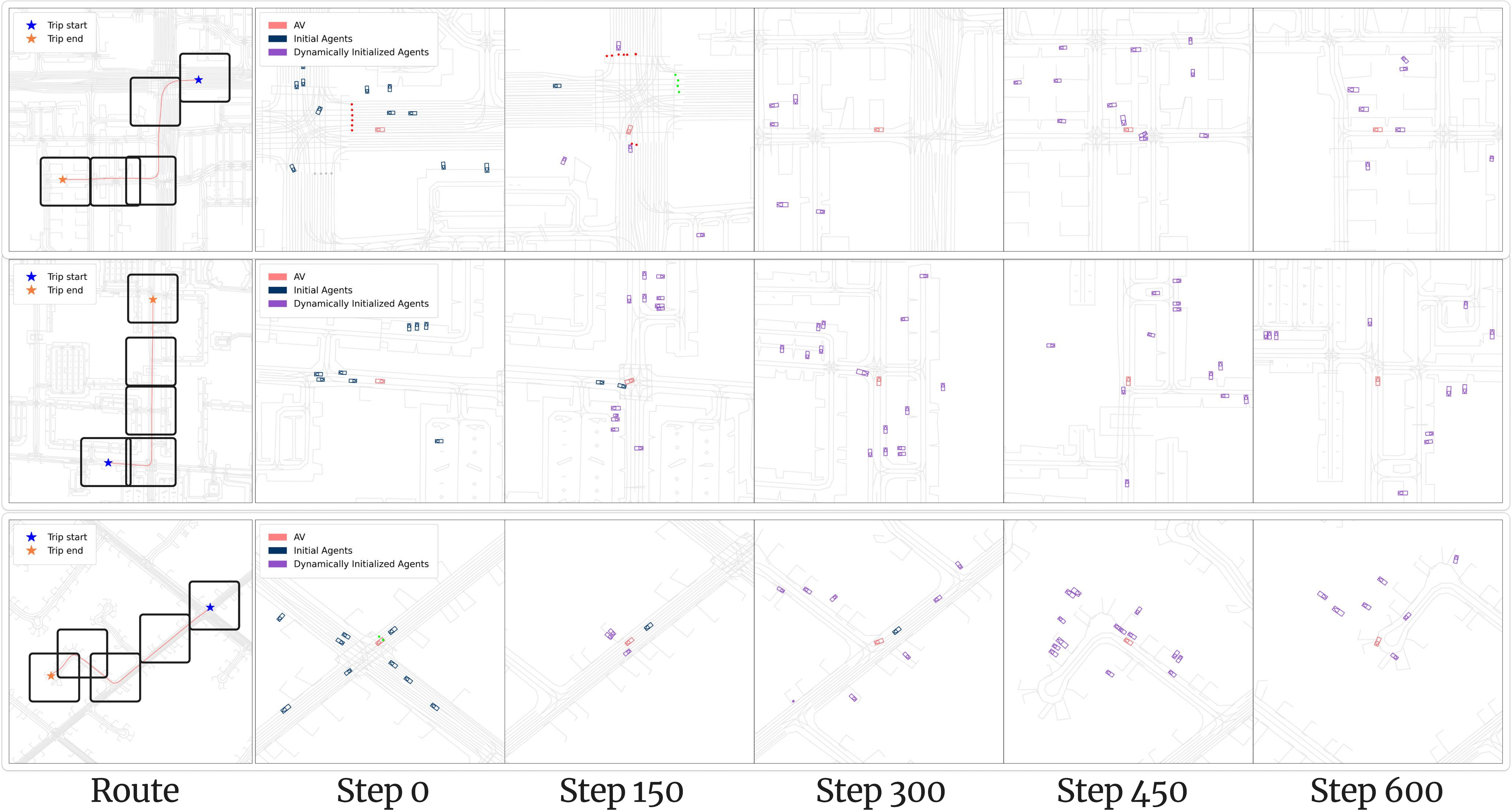}
    \caption{Additional qualitative results of \ours{}. \textit{(From left to right)} Snapshots of full 60s rollouts at 0, 15, 30, 45 and 60-second timesteps.}
    \label{fig:supp-qual2}
\end{figure*}

\begin{figure*}[!h]
    \newcommand{\margin}{12px}
    \includegraphics[trim={\margin, \margin, \margin, \margin},clip,width=1.0\textwidth]{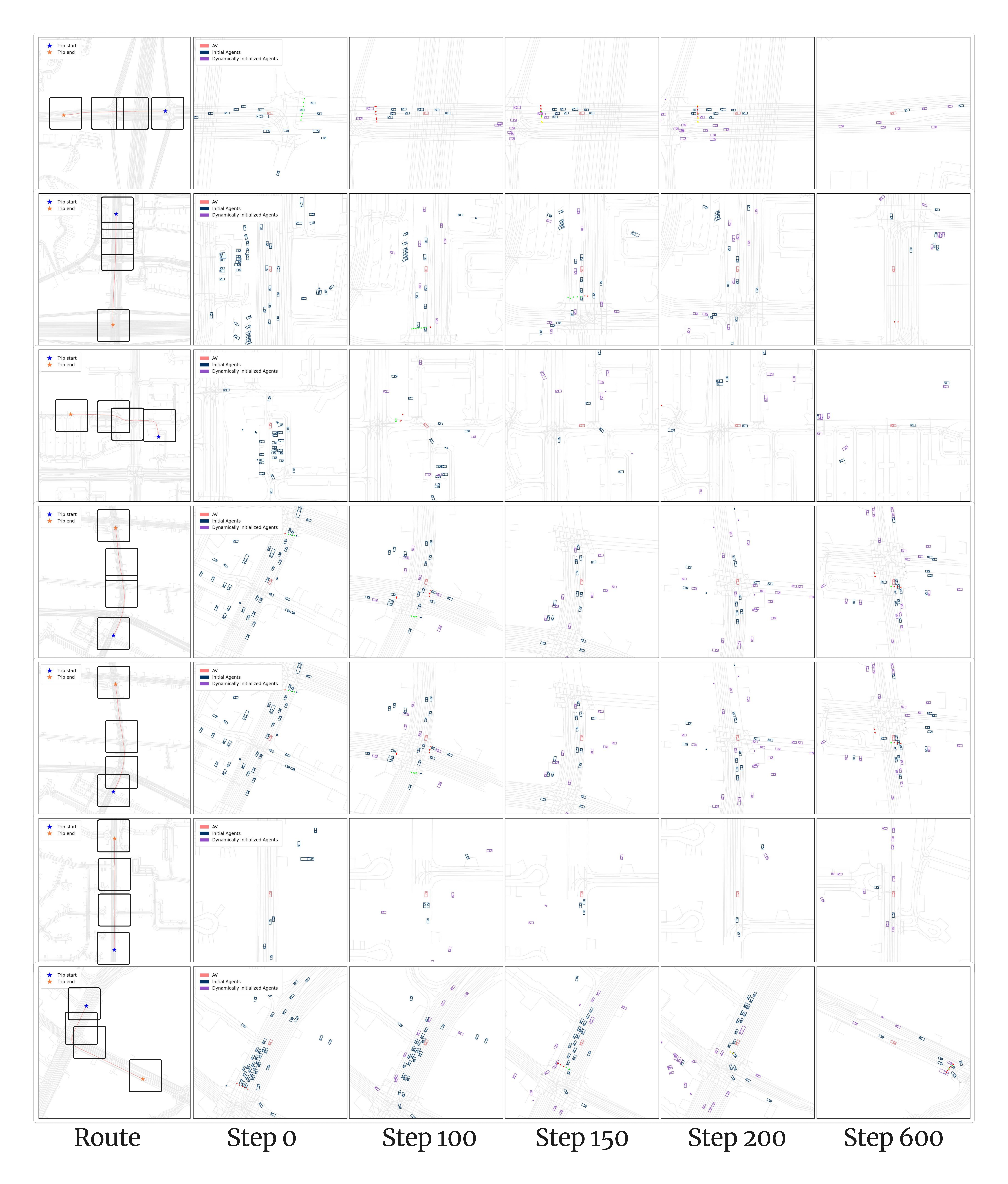}
    \caption{Additional qualitative results of \ours{}. \textit{(From left to right)} Snapshots of full 60s rollouts at 0, 10, 15, 20 and 60-second timesteps.}
    \label{fig:supp-qual1}
\end{figure*}

\begin{itemize}
    \item The city-scale traffic simulation task.
    \item \ours{} architecture and training process.
    \item Rollout videos of \ours{} across long horizons.
\end{itemize}

We encourage readers to watch our video for a better understanding of the long simulation rollout quality from SceneDiffuser++.

\subsection{Model Implementation Details}

\boldparagraph{Architecture}
We use the same context encoder and Transformer denoiser backbone architecture as SceneDiffuser~\cite{Jiang24neurips_scenediffuser}.
Our scene encoder architecture uses 192 latent queries. Each scene token is 512-dimensional, with 8 transformer layers and 8 transformer heads, with a transformer model dimension of 512.
We train and run inference with all 128 agents.

\boldparagraph{Training}
To train \ours, we use the Adafactor optimizer \cite{Shazeer18icml_Adafactor}, with EMA (exponential moving average). We decay using Adam, with $\beta_1=0.9$, $\text{decay}_{adam}=0.9999$, weight decay of 0.01, and clip gradient norms to $1.0$. We use a train batch size of 1024, and train for 1.2M steps. We select the most competitive model based on validation set performance, for which we perform a final evaluation using the test set. We use an initial learning rate of $3 \times 10^{-4}$. We use 32 diffusion sampling steps. When training, we mix the behavior prediction (BP) task with the scene generation task, with probability 0.5. The randomized control mask is applied to both tasks.


\label{subsec:normalization}
\boldparagraph{Feature Normalization} To preprocess features, we use scaling constants of $\frac{1}{80}$ for features $x,y,z$, and compute mean $\mu$ and standard deviation $\sigma$ of features $l, w, h$.

We preprocess each agent feature $f$ to produce normalized feature $f^\prime$ via $f^\prime = \frac{f - \mu_f}{2 * \sigma_f}$, where:
\begin{align}
    \mu_l = 4.5, \quad \mu_w = 2.0, \quad \mu_h = 1.75, \quad \mu_k = 0.5 .
\end{align}
and
\begin{align}
    \sigma_l = 2.5, \quad \sigma_w = 0.8, \quad \sigma_h = 0.6, \quad \sigma_k = 0.5.
\end{align}
We scale by twice the std $\sigma$ values to allow sufficient dynamic range for high feature values for some channels.

We conduct a similar feature normalization process for traffic light features. Specifically, we use the same scaling constants of $\frac{1}{80}$ for features $x,y,z$.
We also convert the traffic light validity and one-shot state features to the range of $[-1, 1]$, similar to what we do for agent validity and type features.

\subsection{Additional Results}
In Figures~\ref{fig:supp-qual2} and \ref{fig:supp-qual1}, we show more qualitative results of \ours{}.
Each row depicts a visualization of a 60-second trip-level rollout of our model.
Within each row, we first show the full trip route overview (in the first column), and then subsequently plot  \ours{}'s predictions at intervals corresponding to 0, 10, 15, 20 and 60 seconds from the start of simulation.

\subsection{Experiment Details}

\textbf{Validity Definition} We define a valid timestep for an agent as whether or not that agent appears in the AV’s detection output at a particular timestep. The Entering (or Exiting) Distance is the distance between an agent and the AV, in meters, at the timestep it appears in the AV’s detection for the first (or last) time.

\boldparagraph{Routing Implementation Details}
SceneDiffuser and SceneDiffuser++ do not use goal-oriented routing; in other words, they do not use or ingest a goal location in any way, shape, or form. Fig. 1 depicts with a star the ``trip end'' point, i.e. the final goal of the ADV, but there is no description of how the model is conditioned with the goal.
This is because the main focus of this work is on the world model, while we assume that the planner can utilize any goal- or route-conditioned model for AV control.
Therefore, in our experiments we also do not define a goal or progress-oriented metrics.


    




When used as a planner, IDM explicitly searches for a valid path for the AV from the start location to a randomly selected goal location with a graph search algorithm on the WOMD lane graph. Similarly, when used as a world model, IDM searches for a path for every other agent.

For \ours{}, when used as a planner, we perform a route-unconditioned rollout in the mapped environment.

This is the same for all other agents when used as a world model.
In this way, agents controlled by any of these three models follow a random path in the mapped environment, making it possible for us to only compare the realism aspect of the world models.

\boldparagraph{Traffic Light Transitions} In order to quantitatively and qualitatively analyze the realism of simulated traffic light state transitions, we construct the traffic light transition probability matrix for \ours{} and compare it against that of the ground-truth logs. We visualize the diagonals in Figure 8 of the main paper, and provide additional details below. Specifically, WOMD\footnote{\url{https://waymo.com/open/data/motion/tfexample/}} \cite{Ettinger21iccv_WOMD} contains 9 different traffic light states: \textit{Unknown}, \textit{Green/Red/Yellow Arrow}, \textit{Solid Green/Red/Yellow}, and \textit{Flashing Red/Yellow}.
Specifically, \textit{Unknown} represents the case when the AV can observe the position of the traffic light, but cannot identify its state due to occlusion. We would like the model to predict only realistic traffic light state transitions, e.g. from \textit{Yellow} to \textit{Red}, but not the other way around.

To compute the transition probability matrix, we count all the consecutive timesteps where the traffic light state changes from one state to a different state, and categorize them based on the starting state and end state, accumulating counts in a $9\times9$ transition matrix.
As we observe that self-transitions from one state to itself are predominant, we removed all the self-transition counts (i.e., the diagonal entries on the transition matrix), and normalize the transition matrix to probabilities.
We obtain the matrix in Figure 8 by computing an average over all scenarios in the validation dataset.
To compute the JS-divergence between the transition probabilities for a quantitative comparison, we directly compute the JS-divergence between the ground-truth transition matrix and that produced by \ours{}.
We observe that \ours{} produces very realistic traffic light transitions.




\end{document}